\documentclass[letterpaper]{article} 
\usepackage[preprint]{aaai2027} 
\usepackage[hyphens]{url} 
\usepackage{graphicx} 
\usepackage{natbib} 
\usepackage{caption} 
\usepackage{amsmath}
\usepackage{amssymb}
\usepackage{booktabs}

\title{DIVE: Embedding Compression via Self-Limiting Gradient Updates}

\author{Dongfang Zhao (dzhao@uw.edu)}
\affiliations{}

\begin{document}
\maketitle

\begin{abstract}
High-dimensional language-model embeddings increase storage and search costs, while supervised compressors can overfit when relevance labels are scarce.  We present DIVE (Dimensionality reduction with Implicit View Ensembles), a residual compression adapter codesigned with a self-limiting hinge loss, geometry distillation, and head-wise NT-Xent over implicit coordinate views.  
The hinge stops updating satisfied ranking constraints, while the dense objectives stabilize the compressed representation; only the first head is retained at inference.  
Under query-disjoint evaluation with two LLM2Vec backbones, five BEIR benchmarks, 128d and 256d outputs, and six baselines, DIVE is the strongest adapter on all five primary benchmarks.  It also outperforms PCA and an autoencoder in comparisons against unsupervised compressors.
\end{abstract}

\section{Introduction}

Large language models (LLMs) have become the standard backbone for dense retrieval, with models such as LLM2Vec~\cite{llm2vec} producing high-dimensional embeddings that achieve state-of-the-art performance on benchmarks like BEIR~\cite{beir}.
These embeddings, however, typically span thousands of dimensions, imposing high storage and query latency costs on large-scale vector search infrastructure even with optimized indices, e.g., FAISS~\cite{faiss} and HNSW~\cite{hnsw_tpami20}.
Compressing these embeddings without sacrificing retrieval quality has become a practical necessity.

One approach is to attach a lightweight adapter to a frozen LLM backbone, following the paradigm of~\citet{pmlr-v97-houlsby19a} and LoRA~\cite{lora}.
Matryoshka Representation Learning~\cite{mrl_nips22} introduced multi-scale nested embeddings; Matryoshka-Adaptor~\cite{mrlAdapter_emnlp24} and Search-Adaptor~\cite{searchAdapter_acl24} extended this to supervised dimensionality reduction via rank losses and similarity preservation; SMEC~\cite{smec_emnlp25} further introduced sequential training stages and adaptive dimension selection.
These methods work well when labeled query-document pairs are abundant, and
they remain strong supervised compression baselines in our evaluation.  A
common property of their published objectives is that they do not explicitly
gate label-driven updates after a ranking or similarity constraint has already
been satisfied~\cite{mrlAdapter_emnlp24,searchAdapter_acl24,smec_emnlp25}.
With scarce relevance judgments, such persistent supervision can keep changing
the adapter even after it has corrected the intended ordering, potentially
disturbing useful pretrained geometry.  This creates a more specific design
question: can a compressor retain the task signal for unsatisfied constraints
while switching off unnecessary label-driven updates?
DIVE addresses this question through an identity-initialized residual
architecture and a sparse--dense training objective.

We propose Dimensionality reduction with Implicit View Ensembles (DIVE), which
is an identity-initialized residual adapter trained by sparse task supervision,
dense self-supervised view regularization, and dense geometry preservation.
Drawing on the margin-based metric learning of~\citet{kw_jmlr09} and~\citet{Schroff_2015_CVPR}, DIVE replaces the standard ranking loss with a hinge-based triplet loss that produces exactly zero gradient once a triplet satisfies the margin constraint.
Unlike objectives such as SimCSE~\cite{simcse_emnlp21} that maintain continuous gradient flow, DIVE's self-limiting mechanism bounds the label-driven component of perturbation.
To compensate for the resulting gradient sparsity, DIVE introduces a head-wise NT-Xent contrastive loss~\cite{tchen_icml20} that treats coordinate blocks of the shared adapted embedding as implicit views.
Inspired by multi-view self-supervised learning but requiring no
task-specific or potentially semantics-altering text augmentation, this
objective provides $H(H-1)$ positive pairs per sample and evaluates
$O(B^2H^2)$ pairwise similarities per batch, independently of triplet
satisfaction.
At inference time, only the first head is retained, keeping storage and retrieval cost identical to a standard single-head adapter.



We extensively evaluate DIVE with two LLM2Vec backbones (Mistral-7B and
Sheared-LLaMA-1.3B), across five BEIR~\cite{beir} benchmarks, with CQADupStack averaged over
its 12 communities.  
We compare against the frozen representation, Matryoshka-Adaptor~\cite{mrlAdapter_emnlp24},
Search-Adaptor~\cite{searchAdapter_acl24}, SMEC~\cite{smec_emnlp25}, PCA~\cite{pca}, and an
autoencoder~\cite{autoencoder}.  Under a query-disjoint protocol with one globally selected
configuration, DIVE is the strongest supervised adapter.

\section{Background and Related Work}

\paragraph{Parameter-Efficient Fine-Tuning (PEFT)}
PEFT methods adapt large pretrained models with few trainable parameters.
Houlsby et al.~\cite{pmlr-v97-houlsby19a} introduced bottleneck adapters; LoRA~\cite{lora} injects low-rank matrices into frozen weights.
Prefix-Tuning~\cite{li-liang-2021-prefix} and Prompt Tuning~\cite{lester-etal-2021-power} prepend learnable continuous vectors to the input.
AdapterFusion~\cite{jp_eacl21} and LST~\cite{ysung_nips22} extend adapters to multi-task and memory-efficient settings, and Visual Prompt Tuning~\cite{mjia_eccv22} applies the paradigm to vision.
These methods address classification or generation objectives without considering retrieval-specific geometric constraints.
DIVE instead combines identity-initialized full-dimensional residual adaptation,
self-limiting retrieval supervision, and geometry preservation.

\paragraph{Matryoshka Representation Learning (MRL) and Embedding Compression}
MRL~\cite{mrl_nips22} encodes multi-scale information in a single embedding, enabling flexible truncation and inspiring extensions to generative~\cite{mdm_iclr24} and vision-language~\cite{mqt_nips24} models.
For retrieval, Matryoshka-Adaptor~\cite{mrlAdapter_emnlp24} and Search-Adaptor~\cite{searchAdapter_acl24} apply lightweight residual adapters to frozen embeddings; SMEC~\cite{smec_emnlp25} adds sequential training and adaptive dimension selection.
Text backbones such as LLM2Vec~\cite{llm2vec} and scalable vector search via FAISS~\cite{faiss}, HNSW~\cite{hnsw_tpami20}, and disk-based indices including DiskANN~\cite{diskann_nips19}, SPANN~\cite{spann_nips21}, and ScaNN~\cite{rguo_icml20} form the typical retrieval stack, evaluated on benchmarks like BEIR~\cite{beir}.
However, supervised adapters can continue optimizing objectives beyond the
target compressed space, increasing the risk of geometry distortion when
labeled data is scarce.
DIVE instead applies a gated hinge loss only at the deployed target
dimension and assigns dense optimization to view and geometry
preservation objectives.

\paragraph{Traditional Dimensionality Reduction}
Classical compression methods such as PCA~\cite{pca}, Product Quantization~\cite{pq_tpami11}, Optimized Product Quantization~\cite{opq_cvpr13}, and Iterative Quantization~\cite{ygong_cvpr11} reduce dimensions based on data statistics without retrieval labels.
DIVE instead performs label-guided semantic compression that preserves
retrieval relevance directly.
We include PCA and autoencoder baselines in the evaluation for a suite-level comparison with unsupervised compression.

\paragraph{Metric Learning and Contrastive Representation Learning}
The hinge loss behind DIVE originates from metric learning: large-margin nearest neighbor~\cite{kw_jmlr09}, FaceNet~\cite{Schroff_2015_CVPR}, and triplet networks~\cite{eh_iclr15}.
Contrastive objectives such as SimCLR~\cite{tchen_icml20}, MoCo~\cite{moco_cvpr20}, and supervised contrastive~\cite{pk_icml20} learn invariances by pulling positives together and pushing negatives apart.
SimCSE~\cite{simcse_emnlp21} and Sentence-BERT~\cite{sentencebert_emnlp19} adapt contrastive training for high-quality text embeddings.
Contrastive objectives remain dense even when a retrieval ordering is already
correct.  DIVE therefore gates only the label-driven triplet term;
its dense objectives enforce implicit-view consistency and preserve
frozen-space geometry.

\section{Methodology}

\subsection{Multi-Head Adapter Architecture}
\label{sec:arch}

The adapter must create several training views without multiplying its
inference cost.  We obtain these views from non-overlapping coordinate blocks
of one shared residual representation.  Consequently, every auxiliary signal
updates the same residual network used by the deployed head.

The architecture is asymmetric by design.  All heads participate during
training, but only head~1 is used for retrieval.  This distinction lets the
model exploit multi-view regularization while storing a single
$k$-dimensional vector per item.

\subsubsection{Identity-Initialized Residual Adaptation}
Let $\phi: \mathcal{X} \to \mathbb{R}^{d}$ be a frozen text encoder that maps documents and queries to $d$-dimensional embeddings $\mathbf{z}$.
Rather than replacing this representation with a randomly initialized projection, DIVE learns a full-dimensional residual correction.
The correction operates in the same coordinate system as the frozen embedding and is defined by
\begin{equation}
	\widetilde{\mathbf{z}}
	= \mathbf{z} + \mathbf{W}_2\,\sigma(\mathbf{W}_1\mathbf{z}+\mathbf{b}_1)+\mathbf{b}_2,
\end{equation}
where $\sigma$ is ReLU, $\mathbf{W}_1\in\mathbb{R}^{r\times d}$, and $\mathbf{W}_2\in\mathbb{R}^{d\times r}$ with hidden width $r=2048$.
We initialize $\mathbf{W}_2$ and $\mathbf{b}_2$ to zero, so that
$\widetilde{\mathbf{z}}=\mathbf{z}$ before training.  This identity statement
applies to the full-dimensional residual representation.  The deployed head
starts from the normalized coordinate slice $\mathbf z[1:k]$, not from the
complete $d$-dimensional retrieval representation; geometry distillation is
therefore what explicitly preserves frozen pairwise similarities after
compression.  Zero initialization removes distortion from a random residual
correction while the training objectives learn the compressed head.

\subsubsection{Multi-Head Projection and Normalization}

The first $kH$ coordinates of $\widetilde{\mathbf{z}}$ are reshaped into $H$ separate head vectors.
We implement this partition as
\begin{equation}
	\operatorname{MultiHead}(\widetilde{\mathbf z})
	= \operatorname{norm}_2\!\left(
	\operatorname{reshape}(\widetilde{\mathbf z}_{1:kH},H,k)
	\right),
\end{equation}
where $\operatorname{reshape}(\cdot,H,k)$ splits the $kH$-dimensional vector into $H$ chunks of dimension $k$, and $\operatorname{norm}_2$ normalizes each chunk independently to unit norm.
The result is a set of $H$ unit-norm vectors $\{\mathbf{z}^{[h]} \in \mathbb{S}^{k-1}\}_{h=1}^H$.

Formally, for input embedding $\mathbf{z}$ and head index $h \in \{1, \ldots, H\}$:
\begin{equation}
\mathbf{z}^{[h]} = \frac{\widetilde{\mathbf{z}}[(h-1)k+1 : hk]}{\|\widetilde{\mathbf{z}}[(h-1)k+1 : hk]\|_2},
\end{equation}
where $\widetilde{\mathbf{z}}[a:b]$ denotes the slice from index $a$ to $b$.
Each head receives a distinct coordinate block, while all heads remain coupled through the shared residual hidden layer and the joint training objective.  The blocks therefore provide non-identical views without requiring text augmentation or separate adapters.

\subsubsection{Asymmetric Train-Inference Protocol}

During training, all $H$ heads participate in the loss computation (detailed in Section~\ref{sec:loss}).
At inference time, we discard heads $2$ through $H$ and use only the first head $\mathbf{z}^{[1]}$ for retrieval:
	\begin{equation}
	f_\theta(\mathbf z)^{[1]}
	= \frac{\widetilde{\mathbf z}[1:k]}
	{\lVert\widetilde{\mathbf z}[1:k]\rVert_2}.
\end{equation}
This fixed rule makes the deployed representation independent of auxiliary-head selection.

This asymmetry has three consequences.  It shifts auxiliary computation to
training rather than retrieval.  It also fixes deployment behavior before any
held-out result is observed.

First, the multi-head architecture regularizes training without adding inference cost.
The contrastive loss aligns views of the same embedding, while the triplet and geometry losses make the first head useful on its own.  Because the residual hidden layer is shared, gradients from auxiliary heads also regularize the transformation used to produce head~1.
Storage and search cost remain identical to a standard single-head adapter.

Second, auxiliary heads increase gradient diversity on small datasets.
When labeled triplets are scarce, the triplet loss alone provides a weak learning signal.
The multi-head contrastive loss supplies dense self-supervised gradients by treating the $H$ heads as $H$ distinct views of each sample, effectively multiplying the gradient signal by a factor of $H(H-1)$ positive pairs per sample.
This supplies additional training signal when labeled triplets are scarce.

Third, inference always uses the first head.
Discarding auxiliary heads avoids output ensembling and manual head selection.
The first head receives the triplet and geometry losses while also participating in multi-head contrastive regularization.
Its role is therefore specified before training rather than chosen from test performance.

For input dimension $d$ and hidden width $r$, the total adapter parameter count is:
\begin{equation}
	\begin{aligned}
		|\theta| & = dr + rd + r + d \\
		         & = 2(4096)(2048) + 2048 + 4096 \\
		         & \approx 16.8\text{M parameters},
	\end{aligned}
\end{equation}
for the primary Mistral-7B configuration; the Sheared-LLaMA-1.3B configuration
uses $d=r=2048$ and therefore has approximately 8.4M parameters.  This count is
independent of $H$ and $k$.  The auxiliary views therefore add no parameters
and no inference-time storage.  Their only additional cost is the training-time
evaluation of the view objective.

\subsection{Training Objective}
\label{sec:loss}

DIVE is trained with a composite loss that balances task-specific correction,
implicit-view learning, and geometry preservation:
\begin{equation}
	\mathcal{L} = \mathcal{L}_{\text{triplet}}
	+ \lambda\mathcal{L}_{\text{view}}
	+ \gamma\mathcal{L}_{\text{geo}}.
\end{equation}
The hinge triplet loss acts only on head~1, the view loss regularizes all
heads, and geometry distillation transfers frozen pairwise similarities to
head~1.  We select $\lambda$ and $\gamma$ once using the unweighted mean
validation score across the four configuration-selection benchmarks and then
hold them fixed for every dataset.

Each term has a distinct optimization role.  The triplet term expresses
label-derived orderings, the view term connects coordinate blocks, and the
geometry term transfers similarities from the frozen representation.  Their
combination prevents dense optimization from depending solely on repeated
label pressure.

\subsubsection{Self-Limiting Triplet Loss}

For a mini-batch of labeled triplets $\mathcal{B} = \{(\mathbf{q}_i, \mathbf{d}_i^+, \mathbf{d}_i^-)\}_{i=1}^B$, we compute a margin-based hinge loss on the first head only:
\begin{equation}
	\mathcal{L}_{\text{triplet}} = \frac{1}{B} \sum_{i=1}^B \max\bigl( 0,\, m - \Delta_i \bigr),
\end{equation}
where the signed similarity margin for triplet $i$ is:
\begin{equation}
	\Delta_i = \mathbf{q}_i^{[1]} \cdot \mathbf{p}_i^{[1]} - \mathbf{q}_i^{[1]} \cdot \mathbf{n}_i^{[1]},
\end{equation}
and $m > 0$ is a hyperparameter controlling the strictness of the ordering
constraint.
Here $\mathbf{q}_i^{[1]}$, $\mathbf{p}_i^{[1]}$, $\mathbf{n}_i^{[1]} \in \mathbb{S}^{k-1}$ denote the first head of the query, positive, and negative embeddings respectively after L2-normalization.
Thus every labeled triplet is evaluated in the representation used
at inference.

The hinge function implements a hard gradient gate: when $\Delta_i \geq m$, the triplet contributes exactly zero to the loss.
We call such triplets \textit{inactive}.
Define the active triplet ratio at epoch $t$ as:
\begin{equation}
	\rho(t) = \frac{1}{|\mathcal{B}|} \sum_{i=1}^{|\mathcal{B}|} \mathbb{I}[\Delta_i < m],
\end{equation}
where $\mathbb{I}[\cdot]$ is the indicator function.

The gradient gate bounds only the displacement induced by the label-driven
triplet term.  Under standard assumptions of Lipschitz continuity and bounded
per-triplet gradients, this expected displacement satisfies:
\begin{equation}
	\mathbb{E}\bigl[\|\Delta_{\mathrm{trip}} \mathbf{z}\|_2\bigr]
	\leq \eta L G \sum_{t=1}^T \rho(t),
\end{equation}
where $\eta$ is learning rate, $L$ is the adapter's Lipschitz constant, $G$ bounds gradient norms, and $T$ is total epochs.
The contrastive and geometry terms remain active and are not covered by this
bound.  Consequently, the direct triplet-induced perturbation scales with the
observed active-ratio sum rather than mechanically with all $T$ epochs.

\subsubsection{Head-Wise Contrastive Loss}

The sparse triplet gradient alone provides weak learning signal when most triplets are inactive.
To compensate, we introduce a dense self-supervised objective that treats the $H$ heads of each sample as positive pairs.
This objective remains defined independently of whether a labeled ordering already satisfies the margin.

For a batch of embeddings $\{\mathbf{z}_i\}_{i=1}^B$, let $\mathbf{z}_i^{[h]} \in \mathbb{S}^{k-1}$ denote the $h$-th head of sample $i$.
Define $\mathcal A_{i,h}=\{(j,g):(j,g)\ne(i,h)\}$ as the set of non-self indices.
We compute the multi-positive NT-Xent loss as
\begin{equation}
	\begin{gathered}
	\ell_{i,h}
	= -\frac{1}{H-1}\sum_{h'\ne h}
	\log\frac{\exp(\mathbf z_i^{[h]}\!\cdot\!\mathbf z_i^{[h']}/\tau)}
	{\displaystyle\sum_{(j,g)\in\mathcal A_{i,h}}
	 \exp(\mathbf z_i^{[h]}\!\cdot\!\mathbf z_j^{[g]}/\tau)},\\
	\mathcal L_{\mathrm{contrast}}
	=\frac{1}{BH}\sum_{i=1}^{B}\sum_{h=1}^{H}\ell_{i,h},
	\end{gathered}
\end{equation}
where $\tau>0$ is the temperature.  Different heads of the same sample are
positives; all heads belonging to other samples are negatives.
This creates $H(H-1)$ ordered positive pairs per sample and $(B-1) \times H$ negative comparisons for each anchor head. Across the batch, the loss therefore evaluates $BH(H-1)+B(B-1)H^2=O(B^2H^2)$ pairwise similarities.

The head-wise contrastive loss can be interpreted as multi-view self-supervised learning, where views are derived from the network's parametric structure rather than data augmentation.
The objective encourages the separate coordinate blocks to encode semantically consistent representations while allowing them to emphasize different subspaces.
Unlike image-based contrastive methods (e.g., SimCLR~\cite{tchen_icml20}) that rely on cropping or color jittering, DIVE generates views by fixed coordinate partitioning of the learned adapter output, avoiding the need for domain-specific text augmentation.

The contrastive loss is applied separately to queries, positives, and negatives:
\begin{equation}
	\begin{aligned}
		\mathcal{L}_{\text{view}} = \frac{1}{3} \bigl(
			 & \mathcal{L}_{\text{contrast}}(\{\mathbf{q}_i\})
			 + \mathcal{L}_{\text{contrast}}(\{\mathbf{p}_i\}) \\
			 & + \mathcal{L}_{\text{contrast}}(\{\mathbf{n}_i\}) \bigr).
	\end{aligned}
\end{equation}
The same view criterion applies to all embedding types and prevents spurious correlations between embedding type and head structure.
It also exposes queries, positives, and negatives to the same view-consistency criterion.
No additional relevance labels are required for this term.

\subsubsection{Geometry Distillation}

The identity initialization anchors the start of optimization; we additionally
preserve geometry throughout training.  Let
$\mathbf{T}\in\mathbb{R}^{3B\times d}$ concatenate the frozen normalized
queries, positives, and negatives in a batch, and let
$\mathbf{S}\in\mathbb{R}^{3B\times k}$ concatenate their normalized first
heads.  We match their off-diagonal Gram matrices:
\begin{equation}
	\mathcal{L}_{\text{geo}} = \frac{1}{|\Omega|}
	\sum_{(i,j)\in\Omega}\operatorname{SmoothL1}\!\left(
	(\mathbf{S}\mathbf{S}^{\top})_{ij},
	(\mathbf{T}\mathbf{T}^{\top})_{ij}\right),
\end{equation}
where $\Omega=\{(i,j):i\ne j\}$.  This objective requires no additional
labels: it preserves the frozen encoder's pairwise structure while the hinge
loss corrects retrieval orderings.

\subsubsection{Gradient Complementarity}

Early in training, all three terms contribute: the hinge loss corrects violated
orderings, NT-Xent couples the implicit views, and geometry distillation
discourages destructive movement away from the frozen space.  As more triplets
satisfy the margin, the hinge component becomes sparse while both
dense objectives remain active.  The globally fixed weights are
$\lambda=0.01$ and $\gamma=10$; neither is tuned per dataset.

\section{Evaluation}

\subsection{Experimental Setup}

\paragraph{Datasets}
We evaluate five publicly available BEIR benchmarks~\cite{beir}: SciFact,
FiQA, ArguAna, Quora, and the complete 12-community CQADupStack benchmark.
Table~\ref{tab:datasets} gives their corpus and query counts.

\begin{table}[!t]
	\centering
	\begin{tabular}{lrr}
		\toprule
		\textbf{Dataset} & \textbf{\# of Corpus} & \textbf{\# of Queries} \\ \midrule
		SciFact          & 5,183                 & 300                    \\
		FiQA             & 57,638                & 648                    \\
		ArguAna          & 8,674                 & 1,401                  \\
		Quora            & 522,931               & 10,000                 \\
		CQADupStack (CQA)      & 457,199               & 13,145                 \\ \bottomrule
	\end{tabular}
	\caption{BEIR~\cite{beir} dataset statistics.}
	\label{tab:datasets}
\end{table}

\paragraph{Backbone and Embeddings}
We use two backbones: Mistral-7B (4096d) and Sheared-LLaMA-1.3B
(2048d).  
Within each comparison, the backbones use shared splits,
training settings, output dimensions, epochs, and batch sizes.

\paragraph{Baselines}
We compare DIVE with six baseline methods: the uncompressed frozen representation,
Matryoshka-Adaptor~\cite{mrlAdapter_emnlp24}, Search-Adaptor~\cite{searchAdapter_acl24},
SMEC~\cite{smec_emnlp25}, PCA~\cite{pca}, and a standard autoencoder~\cite{autoencoder}.
We use matched residual-MLP
implementations with the respective published losses or descriptions of these
baselines.  For Mistral-7B, the supervised adapters use approximately 16.8M
trainable parameters; the matched Sheared-LLaMA-1.3B adapters use approximately
8.4M.  All use 50 training epochs, AdamW, and batch size 128.
DIVE, Matryoshka-Adaptor, and Search-Adaptor use lr=$2\times10^{-4}$; validation-only stability selection fixes SMEC lr=$5\times10^{-6}$ before held-out evaluation.

\paragraph{Metrics}
We report nDCG@10 and Recall@10, following prior compression work
\cite{searchAdapter_acl24,mrlAdapter_emnlp24}; nDCG emphasizes rank position,
and Recall measures relevant-item coverage in the top ten.

\paragraph{Hyperparameter Selection}
We select one global DIVE configuration by mean validation nDCG@10 on SciFact,
FiQA, ArguAna, and Quora, then freeze $m=0.7$, $\lambda=0.01$, $\gamma=10$,
$\tau=0.1$, and $H=4$ for every study, including CQADupStack.  Section~\ref{sec:matched-tuning}
gives each supervised baseline three validation trials; no held-out test query is used for selection.  The learning rates above define the primary five-benchmark comparison.  The audit in Section~\ref{sec:matched-tuning} is a separate validation-based check using validation-selected learning rates for the three supervised baselines.

\begin{table}[!t]
	\centering
	\small
	\begin{tabular}{l c c c c c}
		\toprule
		\textbf{Evaluation} & \textbf{Margin} $m$ & $\lambda$ & $\gamma$ & $\tau$ & $H$ \\
		\midrule
		All benchmarks & 0.7 & 0.01 & 10 & 0.10 & 4 \\
		\bottomrule
	\end{tabular}
	\caption{Validation-selected DIVE configuration.}
	\label{tab:hyperparams}
\end{table}

\begin{figure*}[!t]
	\centering
	\includegraphics[width=\textwidth]{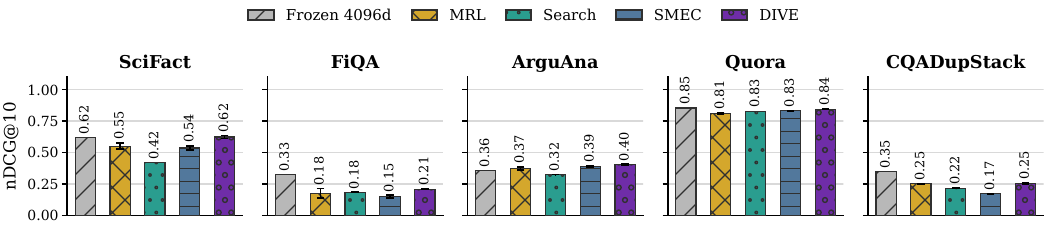}
	\caption{Held-out nDCG@10 at 128d; error bars show one sample standard
		deviation over three fixed seeds.}
	\label{fig:main-ndcg}
\end{figure*}

\begin{figure*}[!t]
	\centering
	\includegraphics[width=\textwidth]{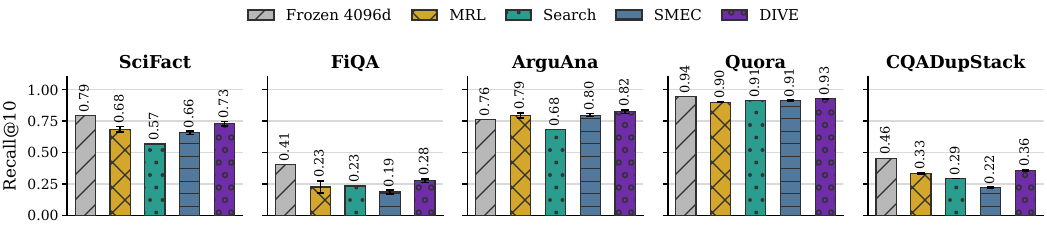}
	\caption{Held-out Recall@10 at 128d under the same protocol and visual
		encoding as Figure~\ref{fig:main-ndcg}.}
	\label{fig:main-recall}
\end{figure*}

\paragraph{Training Protocol}
We partition judged query IDs into disjoint 60\%/20\%/20\% train/validation/test
sets using seed 2027 and SHA-256 query-ID ranks; manifests store the source
fingerprint.  
Selection uses training and validation only; frozen configurations
are retrained on train plus validation and evaluated on held-out queries with
seeds 2027, 2028, and 2029 using shared deterministic triplets.

\paragraph{Platform}
The testing platform uses Ubuntu 24.04 LTS, two AMD EPYC 7763 CPUs, 512GB RAM, and one A100 PCIe
GPU (80GB), with Python 3.12.13, PyTorch 2.5.1/CUDA 12.1, Transformers 4.44.2,
LLM2Vec 0.2.3, BEIR 2.2.0, and \texttt{pytrec\_eval} 0.5.

\subsection{End-to-end Performance}

Figure~\ref{fig:main-ndcg} presents held-out nDCG@10 at 128 dimensions on all five
benchmarks.  Adapter results are mean $\pm$ sample standard deviation over the
three predeclared seeds 2027, 2028, and 2029.  Each CQADupStack entry is first
averaged with equal weight over its 12 communities.

DIVE is the best compressed method on every benchmark.  Relative to the strongest compressed baseline on each dataset, its mean gains are 0.0721 on scifact, 0.0249 on fiqa, 0.0146 on arguana, 0.0119 on quora, and 0.0021 on CQADupStack.  The CQADupStack result uses the same frozen configuration without dataset-specific model selection.

Figure~\ref{fig:main-recall} reports Recall@10 under the same protocol.
DIVE again leads all compressed methods on every benchmark, with an
average Recall@10 of 0.6217.
The Recall@10 gains rule out an explanation based only on rearranging a fixed set of retrieved relevant items.

\subsection{Tuning Fairness and Stability}
\label{sec:matched-tuning}

We conduct a separate learning-rate audit to test whether the primary ranking
is an artifact of an unfavorable baseline learning rate.  MRL, Search-Adaptor,
and SMEC each receive exactly three method-specific learning-rate candidates.
The audit uses the predefined size-controlled subset: SciFact, FiQA, and
ArguAna, the three primary benchmarks with fewer than 100K corpus documents.
We use this fixed subset consistently for experiments requiring independent
retraining and select one learning rate per method by unweighted mean
validation nDCG@10.  
Candidate
grids are method-specific because SMEC's sequential stages operate at a lower
optimization scale, but the number of validation trials is identical.

\begin{figure}[!t]
	\centering
	\includegraphics[width=\columnwidth]{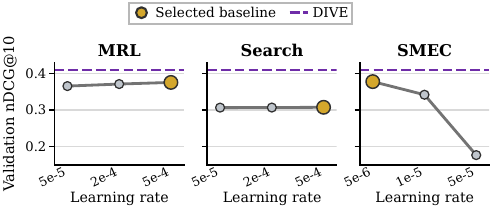}
	\caption{Three-trial baseline; gold
		marks each validation-selected setting and purple shows DIVE.}
	\label{fig:matched-tuning}
\end{figure}

Figure~\ref{fig:matched-tuning} selects learning rates of $5\times10^{-4}$ for
MRL and Search-Adaptor and $5\times10^{-6}$ for SMEC.  After freezing these
choices and retraining on training plus validation queries, DIVE obtains mean
held-out nDCG@10 scores of 0.6223, 0.2087, and 0.4028 on SciFact, FiQA, and
ArguAna.  The strongest tuned baselines reach 0.5592, 0.1837, and 0.3882,
respectively.  DIVE also exceeds the strongest baseline for the corresponding
optimization seed in all nine dataset-seed comparisons.

We next test DIVE's sensitivity to the number of implicit views by varying
$H\in\{2,4,8\}$ while holding every other setting fixed.  Each configuration
is retrained on the same three datasets using seed 2027 and evaluated only on
validation queries.  This diagnostic is not used to alter the globally fixed
$H=4$ configuration employed in the held-out experiments.

\begin{figure}[!t]
	\centering
	\includegraphics[width=\columnwidth]{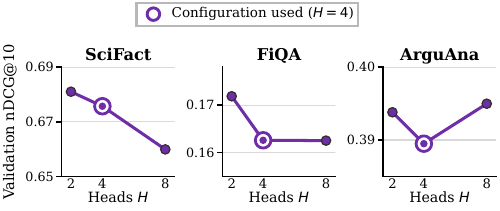}
	\caption{nDCG@10 as the number of heads varies; the
		ring marks the configuration used in held-out experiments.}
	\label{fig:head-sensitivity}
\end{figure}

Figure~\ref{fig:head-sensitivity} reports the individual dataset results rather
than only their average.  SciFact varies from 0.6599 to 0.6810, FiQA from
0.1625 to 0.1719, and ArguAna from 0.3895 to 0.3950 across the tested range.
Moreover, every one of the nine dataset--$H$ configurations exceeds the
strongest tuned baseline on the corresponding validation set.  The best head
count differs by dataset, supporting stability rather than dependence on a
single favorable value of $H$.

\subsection{Ablation Study}
\label{sec:ablation}

We isolate DIVE's core components on the predefined size-controlled subset.  Each
variant requires an independent 50-epoch retraining run, so this size rule
bounds the additional computational cost. 

The \textit{without self-limiting} variant replaces the hinge with a softplus
loss using the same margin, so satisfied triplets continue to produce
gradients.  The other two variants set $\lambda=0$ or $\gamma=0$ to remove
contrast or geometry distillation, respectively.  All other settings are held
fixed and no held-out test query participates in this analysis.

\begin{table}[t]
	\centering
	\small
	\begin{tabular}{l c}
		\toprule
		Configuration & Mean validation nDCG@10 \\
		\midrule
		w/o self-limiting & 0.2793 \\
		w/o contrast & 0.3991 \\
		w/o geometry & 0.3555 \\
		DIVE & \textbf{0.4093} \\
		\bottomrule
	\end{tabular}
	\caption{Unweighted mean validation nDCG@10 at 128d.}
	\label{tab:ablation}
\end{table}

As shown in Table~\ref{tab:ablation}, replacing the self-limiting hinge with an
always-active smooth loss produces the largest mean drop, from 0.4093 to
0.2793.  Removing implicit-view contrast reduces the mean to 0.3991, while
removing geometry distillation reduces it to 0.3555.  Full DIVE therefore
obtains the highest mean among the four configurations.

\subsection{Training Dynamics}
\label{sec:dynamics}

\begin{figure}[t]
	\centering
	\includegraphics[width=\linewidth]{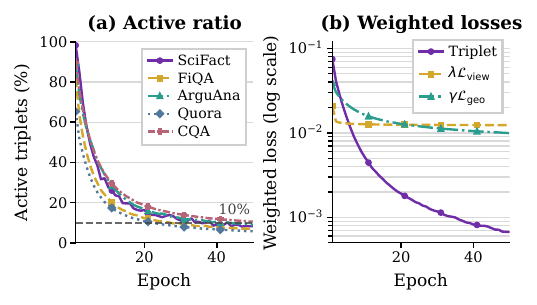}
	\caption{Active-triplet ratios (left), and weighted losses on Quora (right).}
	\label{fig:dynamics}
\end{figure}

Figure~\ref{fig:dynamics} (left) tracks the active triplet ratio $\rho(t)$ throughout training.
Recall that a triplet is \textit{active} at epoch $t$ if its similarity margin $\Delta < m$, meaning that it still contributes non-zero gradient.
The curves therefore measure how much label-driven supervision remains at each epoch.
The active ratio falls below 10\% at epochs 36, 28, 42, and 23 on SciFact, FiQA, ArguAna, and Quora, respectively.  At epoch 50 their ratios are 8.2\%, 6.9\%, 9.0\%, and 5.7\%; the unweighted mean over CQADupStack's 12 communities reaches 10.6\%.  Thus, roughly 90\% or more of the triplets cease receiving label-driven gradients across all five benchmarks despite their different corpus sizes and domains.

The loss decomposition (Figure~\ref{fig:dynamics}, right) shows
$\mathcal{L}_{\text{triplet}}$ decaying as triplets become inactive, while
view regularization and geometry preservation continue to provide dense
training signals.  When labeled supervision is limited, dense objectives
remain active as the triplet loss becomes sparse.  DIVE therefore makes task-driven corrections while constraints are unsatisfied, then relies primarily on
the dense objectives once most ranking constraints are met.  This observed
sparse--dense transition supports the intended separation between label-driven
adaptation and representation stabilization.

\subsection{Multi-Dimensional Results}
\label{app:multidim}

\begin{figure}[!t]
	\centering
	\includegraphics[width=\columnwidth]{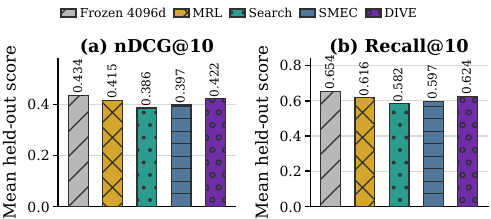}
	\caption{Mean held-out results over three benchmarks at 256d; Frozen is 4096d.}
	\label{fig:256d}
\end{figure}

Figure~\ref{fig:256d} complements the primary 128-dimensional results with a
256-dimensional evaluation.  It reports the unweighted mean over the
predefined size-controlled subset.  
Every method uses the same configuration, seed
2027, and query-disjoint training and evaluation protocol.

At 256 dimensions, DIVE obtains the highest mean scores among compressed
methods: 0.4217 nDCG@10 and 0.6242 Recall@10.  The corresponding MRL scores are
0.4151 and 0.6159, giving DIVE gains of 0.0066 and 0.0083.  Search-Adaptor and
SMEC are lower on both metrics.
The uncompressed backbone remains higher, with mean nDCG@10
of 0.4341 and Recall@10 of 0.6537.  Figure~\ref{fig:256d} therefore measures
quality at a larger compression budget rather than claiming improvement over
the full-dimensional representation.  DIVE remains the strongest compressed
method when the output dimension doubles from 128 to 256.

\subsection{Results across Backbones}
\label{sec:secondary-backbone}

We evaluate the same setting with multiple LLM2Vec backbones.  The
Mistral-7B backbone has 4096-dimensional frozen embeddings, while
Sheared-LLaMA-1.3B has 2048-dimensional frozen embeddings.  For the latter, we
use the predefined size-controlled subset.  We reuse the primary query splits and global
hyperparameters without further tuning for this backbone or these datasets.

\begin{figure}[!t]
	\centering
	\includegraphics[width=\columnwidth]{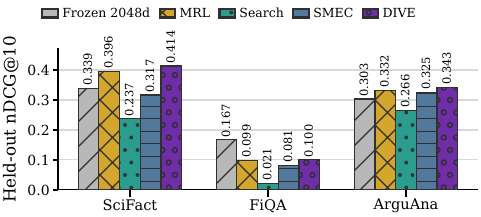}
	\caption{Per-dataset nDCG@10 with LLM2Vec-Sheared-LLaMA-1.3B at 128d; Frozen is 2048d.}
	\label{fig:secondary-backbone}
\end{figure}

In Figure~\ref{fig:secondary-backbone}, DIVE is the strongest compressed
method on all three datasets, reaching nDCG@10 values of 0.4137, 0.0996, and
0.3425 on SciFact, FiQA, and ArguAna, respectively.  The corresponding strongest
compressed baselines obtain 0.3958, 0.0990, and 0.3316.  DIVE's advantage
therefore transfers from the 7B primary backbone to a 1.3B backbone, whose
frozen embeddings are 2048-dimensional rather than 4096-dimensional.

\subsection{Unsupervised Compressors}
\label{app:unsupervised}

Table~\ref{tab:unsupervised} compares DIVE against two unsupervised compression
baselines at 128 dimensions: PCA~\cite{pca} and a standard
autoencoder~\cite{autoencoder}, both trained on frozen LLM2Vec-Mistral-7B
embeddings without retrieval supervision.  We use the predefined
size-controlled subset.

\begin{table}[h]
	\centering
	\small
	\begin{tabular}{l c}
		\toprule
		Method & Mean held-out nDCG@10 \\
		\midrule
		PCA & 0.3284 \\
		Autoencoder & 0.3905 \\
		DIVE & \textbf{0.4113} \\
		\bottomrule
	\end{tabular}
	\caption{Unweighted mean over three benchmarks at 128d.}
	\label{tab:unsupervised}
\end{table}

All methods are evaluated on the same held-out query IDs.  DIVE obtains the
highest three-dataset mean nDCG@10, exceeding the autoencoder by 0.0208 and PCA
by 0.0829.  The comparison isolates retrieval supervision from generic
variance preservation and reconstruction.

\subsection{Retrieval Efficiency}
\label{sec:efficiency}

We benchmark exact top-10 search on all 522,931 Quora vectors using FP32 FAISS
\texttt{GpuIndexFlatIP} on the A100 GPU.  Timing includes query transfer,
search, and result return, but excludes offline embedding generation, file
loading, and index construction.  The 4096d point uses Frozen, whereas 256d
and 128d use DIVE.  We perform five independent runs; each run uses 200, 60,
and 30 timed searches.  

\begin{figure}[t]
	\centering
	\includegraphics[width=\columnwidth]{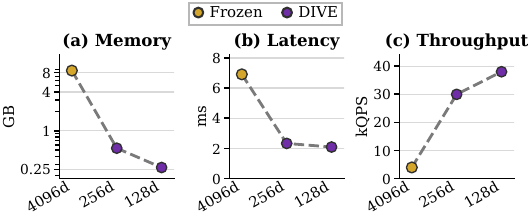}
	\caption{Quora search cost at three embedding dimensions.}
	\label{fig:efficiency}
\end{figure}

Figure~\ref{fig:efficiency} summarizes five independent measurements.  Compared
with the frozen 4096-dimensional index, DIVE's 256- and 128-dimensional indices
use $16\times$ and $32\times$ less memory, respectively.  They also reduce
single-query latency by $2.96\times$ and $3.31\times$, and increase batched
throughput by $7.50\times$ and $9.51\times$.  These efficiency gains come from
using fewer dimensions; DIVE combines them with the strongest compressed
retrieval quality.

\section{Conclusion}

DIVE is an embedding-compression adapter designed to limit unnecessary
label-driven changes to a pretrained representation.  A self-limiting hinge
loss stops updating satisfied ranking constraints, geometry distillation anchors
the compressed space, and head-wise NT-Xent supplies dense gradients through
implicit coordinate views; only the first head is retained at inference.
Across five BEIR benchmarks, two LLM2Vec backbones, 128d and 256d outputs, and
six baselines, DIVE consistently provides the strongest supervised compression
under a query-disjoint evaluation protocol.  It also outperforms unsupervised
PCA and an autoencoder while retaining the memory and latency benefits of
low-dimensional retrieval.  
These results demonstrate
that limiting unnecessary label-driven updates can preserve retrieval quality
while making embedding compression practical.

\bibliography{custom}

\clearpage

\appendix

\section{Theoretical Properties of DIVE}
\label{app:theory}

\subsection{Gradient Sparsity Bound}
\label{app:sparsity}

We derive a bound on the label-driven component of perturbation discussed in
Section~\ref{sec:loss}.  The view and geometry objectives are dense by
construction and are not covered by this sparsity bound.
The resulting statement therefore concerns only the direct gradient signal of the hinge term.

Let the adapter $f_\theta: \mathbb{R}^d \to \mathbb{R}^k$ be $L$-Lipschitz
continuous with respect to its parameters, i.e., for any input $\mathbf{z}$
and any $\theta_1, \theta_2$:
\begin{equation}
	\bigl\lVert f_{\theta_1}(\mathbf z)-f_{\theta_2}(\mathbf z)\bigr\rVert_2
	\le L\lVert\theta_1-\theta_2\rVert_2.
\end{equation}
Assume that the gradient of the triplet loss with respect to the parameters is uniformly bounded: for every triplet $(q,p,n)$ and any parameter setting encountered during training, $\|\nabla_\theta \ell_{q,p,n}(\theta)\|_2 \leq G$.
Let $\mathbf{z}$ be an arbitrary corpus embedding.
For an idealized gradient-flow (or SGD) analysis, let
$d\theta_{\mathrm{trip}}/dt$ denote the parameter-update component induced by
the triplet gradient.  The corresponding cumulative output displacement after
$T$ epochs is bounded by:
\begin{equation}
	\Delta_{\mathrm{trip}}(\mathbf z)
	=\left\lVert\int_0^T \nabla_\theta f_\theta(\mathbf z)
	\frac{d\theta_{\mathrm{trip}}}{dt}\,dt\right\rVert_2.
\end{equation}

By the triangle inequality for integrals and the sub-multiplicativity of the
spectral norm:
\begin{equation}
	\Delta_{\mathrm{trip}}(\mathbf z)
	\le L\int_0^T\left\lVert
	\frac{d\theta_{\mathrm{trip}}}{dt}\right\rVert_2dt.
\end{equation}
This step isolates the length of the triplet-induced parameter trajectory.
It converts output displacement into a quantity controlled by the active
gradient support.

This decomposition describes the direct gradient signal of the triplet term;
it is not an exact decomposition of AdamW's moment-normalized update.
Only active triplets contribute a non-zero gradient; the fraction of such triplets is $\rho(t)$.
Since inactive triplets contribute exactly zero gradient, the expected norm of
the mini-batch gradient satisfies
\begin{equation}
	\mathbb E\!\left[\left\lVert\frac{1}{|B|}
	\sum_{\tau\in B}\nabla_\theta\ell_\tau\right\rVert_2\right]
	\le G\rho(t).
\end{equation}
Hence $\mathbb E[\lVert d\theta_{\mathrm{trip}}/dt\rVert_2]
\le\eta G\rho(t)$, where $\eta$ is the learning rate.
Taking expectations and inserting this bound:
\begin{equation}
	\mathbb{E}\bigl[\Delta_{\mathrm{trip}}(\mathbf{z})\bigr] \leq \eta L G \int_0^T \rho(t) \, dt.
\end{equation}

The empirical observations in Section~\ref{sec:dynamics} show that $\rho(t)$
declines during training and approaches a smaller late-stage value $\rho^*$.
We model this decay as:
\begin{equation}
	\rho(t) \leq \rho_0 e^{-\beta t} + \rho^*,
\end{equation}
where $\rho_0$ is the transient amplitude, $\beta > 0$ controls the
decay rate, and $\rho^* \ll 1$ is the asymptotic active ratio.
Under this model the integral evaluates to:
\begin{equation}
	\int_0^T \rho(t) \, dt \leq \frac{\rho_0}{\beta}(1 - e^{-\beta T})
	+ \rho^* T \leq \frac{\rho_0}{\beta} + \rho^* T.
\end{equation}

Substituting gives the bound:
\begin{equation}
	\mathbb{E}\bigl[\Delta_{\mathrm{trip}}(\mathbf{z})\bigr] \leq \eta L G
	\left(\frac{\rho_0}{\beta} + \rho^* T\right).
	\label{eq:bound}
\end{equation}
This expression isolates the dependence on the transient and residual active
ratios.
Across the four primary tasks, the epoch-50 active ratios range from $0.057$
on quora to $0.090$ on arguana.  Even the largest measured value makes the instantaneous
bound $0.090\,\eta LG$, compared with $\eta LG$ when every triplet remains
active, an $11.1\times$ reduction in the label-driven component.

\subsection{Interpretation of the Perturbation Bound}
\label{app:comparison}

The bound distinguishes gradient gating from a persistently active supervision
setting.  It does not assume that a named baseline necessarily realizes either
extreme.  Instead, the two cases provide reference trajectories for the
label-driven component.

\begin{table}[h]
	\centering
	\small
	\begin{tabular}{lcc}
		\toprule
		\textbf{Case} & \textbf{$\rho(t)$} & \textbf{Triplet bound} \\
		\midrule
		Always active & $1$ & $O(T)$ \\
		Self-limiting & $\rho(t) \to \rho^*$ &
		$O\!\left(\frac{\rho_0}{\beta} + \rho^* T\right)$           \\
		\bottomrule
	\end{tabular}
	\caption{Conditional comparison of label-driven perturbation growth.}
	\label{tab:perturbation}
\end{table}

Table~\ref{tab:perturbation} contrasts the two limiting cases without
assuming that any particular baseline is always active.  An ungated or
persistently active label objective yields linear growth in this bound;
DIVE's observed decay instead scales the triplet-induced component by
the active-ratio trajectory.  Identity initialization and geometry
distillation address the separate question of where dense optimization
starts and what structure it preserves.

This comparison is conditional on the bounded-gradient and Lipschitz
assumptions stated above.  It characterizes direct triplet-induced movement,
not the complete displacement produced by AdamW and the dense objectives.
Accordingly, the empirical active-ratio curves complement rather than prove
the bound.

\subsection{Gradient Diversity of Multi-Head Contrastive Learning}
\label{app:multihead}

We analyze why the multi-head contrastive objective provides richer gradient signal than a single-head variant, formalizing the gradient diversity argument from Section~\ref{sec:arch}.
The relevant quantity is the number of within-sample positives and cross-sample negatives exposed to the shared adapter.
This count describes available supervision without assuming that all pair gradients are independent.

Let $\{\mathbf{z}_i^{[h]}\}_{h=1}^H$ denote the $H$ head vectors produced by the adapter for sample $i$.
The NT-Xent loss for a batch of size $B$ creates $H(H-1)$ positive pairs per sample (one for each ordered pair of distinct heads) and $(B-1) \times H$ negative pairs per head.
The total number of comparison pairs per batch is therefore:
\begin{equation}
	\begin{aligned}
	N_{\mathrm{pairs}}
	&=BH(H-1)+B(B-1)H^2\\
	&=BH\bigl(H-1+(B-1)H\bigr).
	\end{aligned}
\end{equation}

For a single-head variant ($H=1$), the contrastive objective is unavailable
because there are no within-sample positive pairs; the triplet and geometry
objectives remain.
For $H=4$ and $B=128$, $N_{\text{pairs}} = 128 \times 4 \times (3 + 127 \times 4) = 128 \times 4 \times 511 = 261{,}632$ comparison pairs per batch.
Thus the auxiliary heads give each batch a denser comparison set.

Let $\mathbf{g}^{[h]} = \nabla_\theta \mathcal{L}_{\text{contrast}}^{[h]}$ denote the gradient contribution of head $h$.
The total contrastive gradient is $\mathbf{g} = \sum_{h=1}^H \mathbf{g}^{[h]}$.
If the head gradients are sufficiently diverse (non-collinear), the expected
squared norm of the total gradient satisfies:
\begin{equation}
	\begin{aligned}
	\mathbb E[\lVert\mathbf g\rVert_2^2]
	={}&\sum_{h=1}^H\mathbb E[\lVert\mathbf g^{[h]}\rVert_2^2]\\
	&+2\sum_{h<h'}\mathbb E[
	\mathbf g^{[h]}\!\cdot\!\mathbf g^{[h']}].
	\end{aligned}
\end{equation}

Independent normalization does not by itself guarantee orthogonal gradients.
The concrete benefit is instead combinatorial: head-wise NT-Xent supplies
$H(H-1)$ ordered positive pairs per sample, and their gradients meet in the
shared residual hidden layer.  Thus auxiliary blocks can regularize head~1
without assuming that their coordinate subsets overlap.

\subsection{Sparse-Dense Gradient Decomposition}
\label{app:complementarity}

We formalize the sparse-dense gradient decomposition described in Section~\ref{sec:loss}.
The decomposition concerns gradient support before optimizer-specific moment normalization.
It identifies which objectives can become inactive through the hinge gate.

At epoch $t$, the total gradient driving the parameter update is:
\begin{equation}
	\begin{aligned}
	\nabla_\theta\mathcal L(t)
	={}&\underbrace{\nabla_\theta\mathcal L_{\mathrm{triplet}}(t)}_{\text{sparse}}\\
	&+\lambda\underbrace{\nabla_\theta\mathcal L_{\mathrm{view}}(t)}_{\text{dense}}
	+\gamma\underbrace{\nabla_\theta\mathcal L_{\mathrm{geo}}(t)}_{\text{dense}}.
	\end{aligned}
\end{equation}
Only the first term contains a hinge indicator.  The remaining two terms are
weighted by fixed global coefficients.  Their gradients can vanish at an
optimum, but their support is not controlled by $\rho(t)$.

The triplet gradient at epoch $t$ is:
\begin{equation}
	\begin{aligned}
	\nabla_\theta\mathcal L_{\mathrm{triplet}}(t)
	=\frac{1}{B}\sum_{i=1}^B
	&\mathbb I[\Delta_i<m]\\[-2pt]
	&\cdot\nabla_\theta(m-\Delta_i),
	\end{aligned}
\end{equation}
where $\mathbb{I}[\Delta_i < m]$ is the hard gate from the hinge function.
The effective number of contributing triplets is $\rho(t) \cdot B$, which decays rapidly as shown in Section~\ref{sec:dynamics}.
If all constraints are eventually satisfied, then
$\nabla_\theta \mathcal{L}_{\text{triplet}}\to\mathbf{0}$; in finite
training, its expected support is proportional to $\rho(t)B$.

For one set of embeddings, the contrastive gradient at epoch $t$ is:
\begin{equation}
	\nabla_\theta\mathcal L_{\mathrm{contrast}}(t)
	=\frac{1}{BH}\sum_{i=1}^B\sum_{h=1}^H
	\nabla_\theta\ell_{i,h}(t).
\end{equation}
The full view gradient averages this expression over queries, positives, and
negatives, as defined in the main text.
Unlike the triplet gradient, neither dense objective has a hinge gate.
The view term covers every head, while geometry distillation covers every
off-diagonal pair until the corresponding structure is matched.
Their support therefore does not shrink with the active-triplet ratio.

Define the \textit{effective gradient signal} at epoch $t$ as the expected
squared norm of the total gradient:
\begin{equation}
	S(t) = \mathbb{E}\left[\|\nabla_\theta \mathcal{L}(t)\|_2^2\right].
\end{equation}
This scalar summarizes gradient magnitude across mini-batches.  It is used only
to compare early and late training stages.  The approximation below does not
model cross-component alignment.

In early epochs ($\rho(t) \approx \rho_0$), all components contribute.  Ignoring cross terms for compactness:
\begin{equation}
	\begin{aligned}
	S(t)\approx{}&\mathbb E\!\left[\lVert\nabla_\theta
	\mathcal L_{\mathrm{triplet}}\rVert_2^2\right]\\
	&+\lambda^2\mathbb E\!\left[\lVert\nabla_\theta
	\mathcal L_{\mathrm{view}}\rVert_2^2\right]\\
	&+\gamma^2\mathbb E\!\left[\lVert\nabla_\theta
	\mathcal L_{\mathrm{geo}}\rVert_2^2\right].
	\end{aligned}
\end{equation}
The expression retains all three squared component norms.  It omits cross terms
only to expose their separate scaling.  Early training can therefore combine
substantial ranking and dense regularization signals.

In later epochs ($\rho(t) \approx \rho^*\ll 1$), the triplet gradient becomes
sparse; when focusing on the remaining dense signal:
\begin{equation}
	\begin{aligned}
	S(t)\approx{}&\lambda^2\mathbb E\!\left[\lVert\nabla_\theta
	\mathcal L_{\mathrm{view}}\rVert_2^2\right]\\
	&+\gamma^2\mathbb E\!\left[\lVert\nabla_\theta
	\mathcal L_{\mathrm{geo}}\rVert_2^2\right].
	\end{aligned}
\end{equation}
Thus satisfying labeled constraints does not automatically terminate all
learning: the two dense objectives continue to provide signal until their
respective targets are satisfied.
The approximation predicts a transition in which dense regularization becomes
the dominant remaining signal.  Figure~\ref{fig:dynamics} exhibits this
qualitative pattern on the final checkpoints.

\subsection{Margin and Geometric Separability}
\label{app:margin}

The margin $m$ in DIVE's triplet loss has a direct geometric interpretation.
Consider the set of training-triplet margins at a particular parameter state,
and let:
\begin{equation}
	\Delta_{\min}=\min_i\Delta_i,
	\qquad
	\Delta_{\max}=\max_i\Delta_i.
\end{equation}
Three cases arise depending on the relationship between $m$ and these extrema.
The cases determine whether the hinge gate is closed for all, none, or only a
subset of the observed constraints.

\begin{itemize}
	\item $m \geq \Delta_{\max}$:
	      No triplet can strictly exceed the required margin; almost all
	      constraints remain active, producing persistent label pressure.

	\item $m \leq \Delta_{\min}$:
	      Every triplet already satisfies the constraint, so the triplet
	      gradient is zero.

	\item $\Delta_{\min}<m<\Delta_{\max}$:
	      The practically relevant case.
	      The adapter intervenes only on the subset with $\Delta_i<m$.
\end{itemize}

The globally fixed margin $m=0.7$, which is selected by the unweighted mean validation
score across the four configuration-selection benchmarks rather than
per-dataset test performance, lies in the third case.
It therefore gates a proper subset of triplets at the parameter states observed during training.
The active subset changes with optimization, which explains why $\rho(t)$ is a dynamic rather than a dataset constant.

\end{document}